\documentclass[10pt,twocolumn,letterpaper]{article}

 \usepackage[pagenumbers]{cvpr} 

\usepackage{tikz}
\usetikzlibrary{positioning, arrows.meta, shapes.misc} 

\definecolor{cvprblue}{rgb}{0.21,0.49,0.74}
\usepackage[pagebackref,breaklinks,colorlinks,allcolors=cvprblue]{hyperref}

\def\paperID{16543} 
\def\confName{CVPR}
\def\confYear{2026}

\title{Mapping Networks}

\author{
Lord Sen \qquad Shyamapada Mukherjee\\
National Institute of Technology Rourkela\\
Odisha, India\\
{\tt\small lordsen3008@gmail.com}\\
{\tt\small mukherjees@nitrkl.ac.in}
}

\begin{document}
\maketitle
\begin{abstract}
The escalating parameter counts in modern deep learning models pose a fundamental challenge to efficient training and resolution of overfitting. We address this by introducing the \emph{Mapping Networks} which replace the high dimensional weight space by a compact, trainable latent vector based on the hypothesis that the trained parameters of large networks reside on smooth, low-dimensional manifolds. Henceforth, the Mapping Theorem enforced by a dedicated Mapping Loss, shows the existence of a mapping from this latent space to the target weight space both theoretically and in practice. Mapping Networks significantly reduce overfitting and achieve comparable to better performance than target network across complex vision and sequence tasks, including Image Classification, Deepfake Detection etc, with $\mathbf{99.5\%}$, i.e., around $500\times$ reduction in trainable parameters.

\end{abstract} 
\section{Introduction}
\label{sec:intro}

The universal approximation theorem (\emph{UAT}) states that neural networks with a certain structure can, in principle, approximate any continuous function to any desired degree of accuracy. Today deep learning is characterized not only by big data but also by big models, with trainable parameters ranging from millions to trillions. The training of such a network $f$, with inputs ${x}_i$ and parameters $\vec{\theta}$, is optimization of the loss function,
\begin{equation}
    \frac{1}{N}\sum_{i=1}^N\mathcal{L}(f_{\vec{\theta}} ({x}_i), y_i),
\end{equation} via gradient descent to update $\vec{\theta}$. As these networks grow in size and complexity, it's training becomes more challenging, computation intensive, time consuming and costly.
This difficulty stems from two main issues: (1) training these networks from scratch, and (2) the complexity of optimizing and tracking such huge parameter spaces, which often hinders explainability and leads to overfitting.

This necessitates the search for more efficient training methods. In order to improve the training, we have mainly two choices: (1) Decrease the training time, (2) Decrease the trainable parameters. The first can be addressed by distributed training across multiple GPUs. The second is also important as it reduces the chances of overfitting, increases generalization and reduces the black box nature of the model.

\begin{figure}[t]
  \centering
   \includegraphics[width=0.98\linewidth]{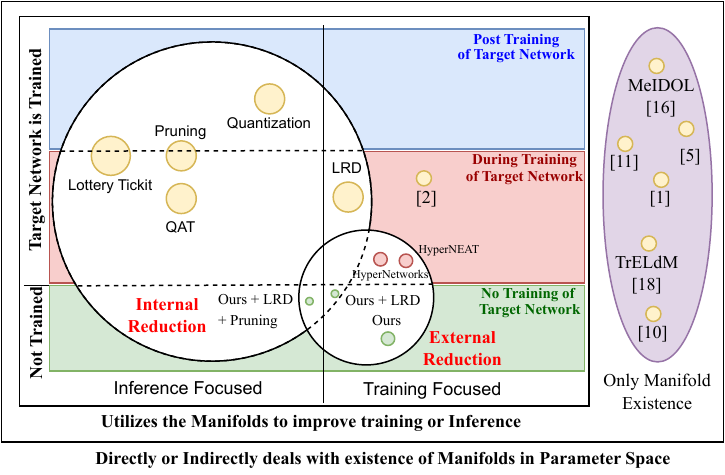}
   \caption{State of the Existing Works and Ours in this field }
   \label{fig:relwor}
\end{figure}

In this context, manifold hypothesis \cite{fefferman2016testing}, \cite{narayanan2010sample} solves the big data problem by assuming high dimensional data to lie on or near a low dimensional manifold. Mathematically, for $x \in  \mathcal{X}$ the high dimensional input space in $\mathbb{R}^D$, $\exists$ $\mathcal{M}$, a low dimensional manifold, such that, $\mathcal{M} \subset \mathcal{X}$ with dimension $d = \dim(\mathcal{M}) \ll D$. 
Then by manifold hypothesis neural networks learn a function,
\begin{equation}
    f_{\theta}: \mathcal{M} \rightarrow \mathcal{Y}.
\end{equation}
In this regard, we have analyzed the parameter space in search of existence of manifolds, around which the trained parameters might be centered. The theory behind our approach is deeply rooted in the study of the geometry of the parameter space. Research on the shape of the loss landscape shows that network weights effectively lie within low-dimensional regions, i.e., have less intrinsic dimension \cite{li2018measuring}.
Supporting this, detailed second-order studies of deep linear networks \cite{achour2024loss}  have found important low-dimensional valleys in the error function. This is further supported by empirical studies demonstrating that the training trajectories of diverse deep networks converge and reside within a shared, intrinsic low-dimensional manifold \cite{mao2024training}, same is supported by our Fig. \ref{fig:pca_submanifold}, \ref{fig:tsne_submanifold}. Similar, conclusions can be drawn from \cite{garipov2018loss}, \cite{draxler2018essentially}, \cite{frankle2020linear} which suggests the existence of manifolds. This is utilized to direct the training trajectory in large flat regions of the energy landscape in \cite{chaudhari2019entropy}. Various inference weight reduction techniques like Pruning \cite{reed1993pruning}, Lottery Ticket hypothesis~\cite{frankle2018lottery} effectively exploit a similar idea as in Figure~\ref{fig:relwor}.

Other techniques such as Low-Rank Compression \cite{idelbayev2020low} reduce redundancy by applying post-training factorization (like SVD) or imposing algebraic constraints $W \approx U V^\top$ on individual weight matrices \cite{eckart1936approximation}. However, these methods operate directly on the high-dimensional weight tensors, either through external compression or a priori linear constraints. A method of predicting the remaining weight values from few given weights of each feature is presented in \cite{denil2013predicting}. Other methods for reducing parameters in convolutional networks are shown in \cite{yang2015deep}, \cite{fernando2016convolution}, \cite{jaderberg2014speeding}.

Our approach is fundamentally different; it is a meta-parametrization. We learn a non-linear, differentiable map $g: \mathbb{R}^d \to \mathbb{R}^P$ that generates the weights from a compact latent vector. This simplifies the problem domain from the high-dimensional weight space to the low-dimensional latent space, which inherently constrains the search to a structurally efficient manifold. This architectural choice naturally promotes the discovery of flatter, more robust solutions in the effective parameter space, providing a structural guarantee of efficiency and stability, placing us in an ideal spot in Figure~\ref{fig:relwor} with respect to training.

In architectural terms, our Mapping Network is a type of Hypernetwork \cite{ha2017hypernetworks}, a model that generates the weights of a target network, which we term as external reduction in Figure~\ref{fig:relwor}. But in HyperNetworks both the target network and the hyper network are trained together, hence the training of the target network cannot be avoided, but not in ours. Also, they do not achieve the same degree of trainable parameter reduction as we do. Modern hypernetworks \cite{gonzalez2023scale} often focus on generating conditional weights for image analysis tasks, enabling rapid adaptation across inputs or tasks but sometimes face issues with stability and guaranteed capacity. We explicitly engineer our system to satisfy the analytical requirements of the Mapping Theorem, which is reflected in our Mapping Loss ($\mathcal{L}_{\text{map}}$) to solve this issue.

Keeping in mind the constraints of our computational resources, we tested our Mapping Networks on modern CNNs and LSTMs, showcasing results on tasks like image classification, image segmentation, and deepfake detection, time series predicton on datasets including Celeb-DF, FF++, MNIST, FMNIST, Cityscapes etc. while maintaining significantly fewer trainable parameters, than baseline. But, the approach being baseline agnostic, can be extended to larger models and datasets as well. 

The main contributions of this work are:
\begin{itemize}
    \item The Mapping Theorem, which establishes the existence of a smooth, low-dimensional parameterization, capable of generating the optimal high-dimensional weights for a target network with an arbitrarily small bounded error.
    \item The Mapping Network, consisting a low-dimensional trainable latent Vector and modulated Mapping Weights to efficiently produce target network parameters, hence decoupling training from target network.
    \item The Mapping Loss, which jointly optimizes task performance and enforces the geometric and analytic properties required by the Mapping Theorem. 
\end{itemize}   
\section{Methodology}
\label{sec:Method}

Inspired by the Manifold Hypothesis discussed in Section~\ref{sec:intro}, we analyze the parameter space in search of existence of manifolds, around which the trained parameters might be centered. The evolution of neural network parameters during training can be interpreted through the lens of differential geometry as the trajectory of points on a low-dimensional manifold embedded in a high-dimensional parameter space $\mathbb{R}^P$. To empirically examine this hypothesis, we recorded parameter snapshots of each layer of a small convolutional neural network (CNN) trained on MNIST.

\begin{figure}[ht]
    \centering
    \begin{subfigure}[b]{0.23\textwidth}
        \centering
        \includegraphics[width=\textwidth, height=100pt]{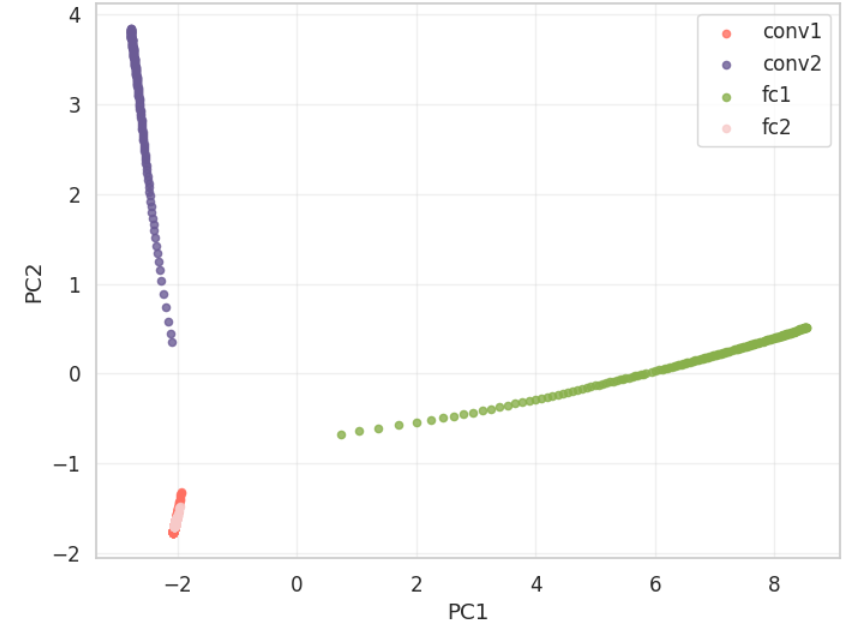}
        \caption{PCA plot.}
        \label{fig:pca_submanifold}
    \end{subfigure}
    \hfill
    \begin{subfigure}[b]{0.24\textwidth}
        \centering
        \includegraphics[width=\textwidth, height=100pt]{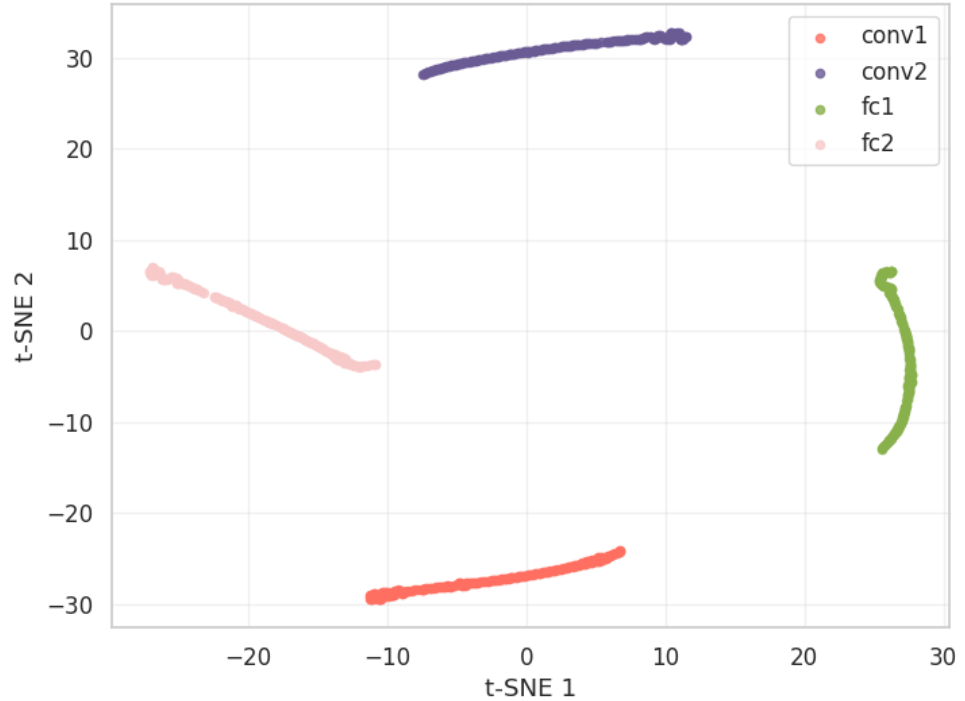}
        \caption{t-SNE plot.}
        \label{fig:tsne_submanifold}
    \end{subfigure}
    
    \caption{Parameter update snapshots showing distinct parameter manifolds in CNN evolution.}
    \label{fig:pca_submanifold22}
\end{figure}
The PCA projection in Figure~\ref{fig:pca_submanifold} reveals that the parameters corresponding to each layer occupy distinct, smooth, and low-dimensional regions in the reduced space. The near-linear trajectories within these regions indicate that, locally, each layer's parameters evolve along approximately affine subspaces. Similarly, the t-SNE plot (Figure~\ref{fig:tsne_submanifold}) highlights the nonlinear geometry of these trajectories.
These observations support the existence of locally Euclidean neighborhoods---a necessary condition for a differentiable manifold structure. This indicates that, during optimization, the parameters do not explore the full $P$-dimensional Euclidean space instead evolve along a smooth, low-dimensional surface. 
Which leads us to state the weight manifold hypothesis as:


\paragraph{Weight–Manifold Hypothesis:  }
For the set of neural parameters $\boldsymbol{\theta} \in \mathbb{R}^P$ of a network $f_{\boldsymbol{\theta}}$, there exists a differentiable embedded manifold $\mathcal{M}_{\boldsymbol{\theta}} \subset \mathbb{R}^P,$ such that $d=\dim(\mathcal{M}_{\boldsymbol{\theta}}) \ll P$ and the trained parameters $\boldsymbol{\theta}^\ast$ lie on (or near) this manifold, meaning all the $P$ values in $\theta^*$ are not independent of each other, i.e., $\boldsymbol{\theta}^\ast \in \mathcal{M}_{\boldsymbol{\theta}}$.

For the CNN, the plots showing smooth and disjoint evolution of each layer’s parameters suggests the existence of differentiable embedded sub manifolds layer wise, i.e., $\boldsymbol{\theta}^{*(l)} \in \mathcal{M}^{(l)}_\theta$.  
This formulation suggests that optimization dynamics implicitly constrain the evolution of parameters to a low-dimensional subspace of $\mathbb{R}^P$, corresponding to a smooth manifold $\mathcal{M}_{\boldsymbol{\theta}}$. 
The existence of such a differentiable manifold implies the possibility of existence of a differentiable mapping 
\begin{equation}
    g : \mathcal{U} \to \mathcal{M}_{\boldsymbol{\theta}} \subset \mathbb{R}^P, \quad \text{where } \mathcal{U} \subset \mathbb{R}^d, \quad d<<P
\end{equation}

from a low dimensional space to the parameter space, which is presented in Section~\ref{ref:mappingtheorem}.

\subsection{Mapping Theorem and Practical Corollary}
\label{ref:mappingtheorem}
Based on the Hypothesis stated above and under certain assumptions, we will state and prove our mapping theorem.
\noindent
\emph{Assumptions:}

\textbf{A1: Smoothness in Parameters:  }The map $\theta \rightarrow f_{\theta}(x)$ is $L_{\theta}$-Lipschitz in parameter form for every $x \in \mathcal{X}$:
\begin{equation}
    \lVert f_{\theta_1}(x) - f_{\theta_2}(x) \rVert \leq L_{\theta} \lVert \theta_1 - \theta_2 \rVert_2, \quad \forall x
\end{equation}

\noindent
\textbf{A2: Loss Lipschitz:  }
The loss $\mathcal{L}(\cdot, y)$ is $L_{\ell}$-Lipschitz in its first argument for each $y$. Combining with A1, the loss difference is controlled by parameter distance:
\begin{equation} \label{eq:lipschitz-loss}
    \lvert \mathcal{L}(\theta_1) - \mathcal{L}(\theta_2) \rvert \leq L_{\ell} L_{\theta} \lVert \theta_1 - \theta_2 \rVert_2
\end{equation}

\noindent
\textbf{A3: Local Approximability:  }
The manifold $\mathcal{M}_{\theta}$ is $C^2$ and has bounded curvature.\\

\textbf{Theorem (Mapping Theorem).} 
Let the parameters $\theta \in \mathbb{R}^P$ of a neural network satisfy the \emph{Weight-Manifold Hypothesis}, i.e., the optimal parameters $\theta^*$ lie on a $C^2$ embedded manifold $\mathcal{M}_{\theta}$ of intrinsic dimension $d^*\ll P$. Suppose the loss $\mathcal{L} : \mathbb{R}^P \to \mathbb{R}$ satisfies the local Lipschitz condition, i.e., there exist constants $L_\theta > 0$, $L_\ell > 0$ and a radius $r > 0$ such that for all $\theta_1, \theta_2 \in B(\theta^*, r)$ a sphere centered at $\theta^*$ with radius $r$,  
$|\mathcal{L}(\theta_1) - \mathcal{L}(\theta_2)| \le L_\ell L_\theta \, \|\theta_1 - \theta_2\|.$ Then for every $\varepsilon > 0$ with $\varepsilon \le L_\ell L_\theta r$, there exists,
\begin{itemize}[leftmargin=30pt]
    \item a $\delta > 0$,
    \item an integer $d \ge d^*$,
    \item a $C^2$ map $g : \mathbb{R}^d \to \mathbb{R}^P$, and
    \item a vector $z^* \in \mathbb{R}^d$,
\end{itemize}
\[
\text{such that }\| g(z^*) - \theta^* \| \le \delta
\quad \text{with} \quad
\delta := \frac{\varepsilon}{L_\ell L_\theta},
\]
and therefore
\begin{equation}
    |\mathcal{L}(g(z^*)) - \mathcal{L}(\theta^*)| \le \varepsilon.
\end{equation}

\emph{Proof:  } 
From \emph{weight manifold hypothesis}, we know $\exists$ a $\mathcal{M}_{\theta}$, and because $\mathcal{M}_\theta$ is a $C^2$ embedded manifold and $\theta^* \in \mathcal{M}_\theta$, by definition there exist:
\begin{itemize}[leftmargin=30pt]
    \item an open set $U \subset \mathbb{R}^{d^*}$ with $0 \in U$, and
    \item a $C^2$ diffeomorphism $\varphi : U \to V \subset \mathcal{M}_\theta$,
\end{itemize}
such that $\varphi(0) = \theta^*$ and $\varphi(U) = V$, an open neighborhood of $\theta^*$ inside $\mathcal{M}_\theta$.
Since, $\varphi$ is $C^2$, it is continuous at $0$.

Let the given $\varepsilon > 0$ satisfy $\varepsilon \le L_\ell L_\theta r$, then we define $\delta$ depending upon $\varepsilon$ as,
\begin{equation}\label{eq:delta}
    \delta := \frac{\varepsilon}{L_\ell L_\theta}, \quad0 < \delta \le r.
\end{equation}
Since $\varphi$ is continuous at $0$, the $\epsilon$--$\delta$ definition of continuity gives, for the chosen $\delta > 0$, there exists $\eta > 0$ such that
\begin{equation}
    \|u - 0\| < \eta \quad \Rightarrow \quad \|\varphi(u) - \varphi(0)\| < \delta.
\end{equation}

That is,
\begin{equation}
\|u\| < \eta \quad \Rightarrow \quad \|\varphi(u) - \theta^*\| < \delta.
\end{equation}


Since we need to show the existence of a $C^2$ map $g$, lets pick open sets $U' \subset U$ with $0 \in U'$ and $\overline{U'} \subset U$. 
Let $\psi : \mathbb{R}^{d^*} \to [0,1]$ be a smooth bump function satisfying $\psi \equiv 1$ on $U'$ and $\psi \equiv 0$ outside $U$. 
Let's define
\begin{equation}\label{eq:g_def}
g(u) = \psi(u)\varphi(u) + (1 - \psi(u))\theta^*.
\end{equation}
Then $g \in C^2(\mathbb{R}^{d^*}, \mathbb{R}^P)$, $g(0) = \theta^*$, and $g(u) = \varphi(u)$ in a neighborhood of $0$.

Since, $0 \in U$ and $\varphi(0) = \theta^*$, we have $g(0) = \theta^*$. Furthermore, by (1) we know any $u$ with $\|u\| < \eta$ (and $u \in U$) satisfies $\|g(u) - \theta^*\| < \delta$. Therefore let's choose, 
\begin{equation}
    z^* \in B(0, \eta) \cap U.
\end{equation}
Then by (1),
\begin{equation}
    \|g(z^*) - \theta^*\| = \|\varphi(z^*) - \theta^*\| < \delta.
\end{equation}

Even exact equality might be achieved by taking $z^* = 0$, since, $z^*=0 \in B(0, \eta) \cap U$, yields $g(z^*) = \theta^*$ exactly. 
Now applying the local Lipschitz conditions. Since $\|g(z^*) - \theta^*\| < \delta \le r$, both $g(z^*)$ and $\theta^*$ lie in the ball $B(\theta^*, r)$. Hence,
\begin{equation}
    |\mathcal{L}(g(z^*)) - \mathcal{L}(\theta^*)| \le L_\ell L_\theta \, \|g(z^*) - \theta^*\| < L_\ell L_\theta \, \delta.
\end{equation}

By Equation~\ref{eq:delta} $\varepsilon =L_\ell L_\theta \, \delta$, So,
\begin{equation}
    |\mathcal{L}(g(z^*)) - \mathcal{L}(\theta^*)| < \varepsilon.
\end{equation}

This completes the proof: for the given $\varepsilon$, we exhibited a $\delta >0$, $d = d^*$, a smooth fixed mapping $g$, and a latent $z^*$ satisfying the required bound. Embedding $\mathbb{R}^{d^*}$ into $\mathbb{R}^d$ via $(u,0) \mapsto g(u)$ readily extends the construction to any $d \ge d^*$. Hence proved.\hfill $\Box$\\



\label{sec:theorem-global-extension}

The Mapping Theorem established the existence of a continuous map $g$ that projects a latent variable in a low-dimensional space to a high-dimensional parameter vector with arbitrarily small bounded error. The following theorem proves that the additive modulation of orthogonally initialized mapping networks with fixed weights and trainable latent vector $z$---as used in the experiments constitutes one such $g$.

\label{thm:global-extension}
\textbf{Theorem 2 (Solvability under additive modulation):    }
Let
$
\theta^* \in \mathbb{R}^P,\quad \omega_0 \in \mathbb{R}^W,\quad z_0 \in \mathbb{R}^d
$
denote respectively the target parameter vector, orthogonally initialized weights, and initial latent vector.  
Let $M:\mathbb{R}^d \to \mathbb{R}^W$ be a $C^2$ modulation ($M(z)=Bz$ for our case~\ref{sec:modulation}) modulating the fixed mapping weights as
$\omega(z) = \omega_0 + M(z)$. Then our mapping network is $ g_\omega(z) := g_{\omega(z)}(z) \in \mathbb{R}^P$ is one such $g$ which satisfies Mapping Theorem.
  

To prove with and without gradient descent we divide the statement in two parts:\\
\noindent\textbf{(2.1) Local solvability.}  
There exists \(\varepsilon>0\) such that for a residual,
\begin{equation}\label{eq:residual}
\mathbf r_\theta := \theta^* - g_{\omega_0}(z_0).
\end{equation}
if \(\|\mathbf r_\theta\| \le \varepsilon\),  then $\exists$ \(\Delta z\), a constant $C>0$ such that,
\begin{equation}\label{eq:local-approx}
\|\Delta z\| = O(\|\mathbf r_\theta\|) \text{  and  }
\|g_\omega(z_0+\Delta z)-\theta^*\| \le C\|\mathbf r_\theta\|^2.
\end{equation}
Consequently, by \eqref{eq:lipschitz-loss},
\begin{equation}
    |\mathcal L(g_\omega(z_0+\Delta z)) - \mathcal L(\theta^*)|
\le L_{\ell}L_\theta\, C \|\mathbf r_\theta\|^2.
\end{equation}

\noindent\textbf{(2.2) Global extension.}  
For any prescribed tolerance \(\varepsilon>0\) $\exists$ constants \(C_2,L_\theta,L_{\ell},r>0\) and a latent vector \(z^*\in\mathbb{R}^d\), obtainable by gradient-based optimization under standard local convergence conditions on \(\mathcal L(g_\omega(z))\), such that
\begin{equation}\label{eq:global-approx}
\| g_\omega(z^*) - \theta^* \| \le \delta, \qquad
|\mathcal L(g_\omega(z^*)) - \mathcal L(\theta^*)| \le \varepsilon,
\end{equation}
where \(\delta = \varepsilon/(L_{\ell}L_\theta)\). Moreover, the latent displacement \(\Delta z^* := z^* - z_0\) satisfies the bound
\begin{equation}\label{eq:dz-bound}
\|\Delta z^*\| \le \sqrt{\frac{\delta}{C_2}},
\end{equation}
and this bound holds independently of the residual \(\mathbf r_\theta\) in \eqref{eq:residual}.
Detailed proofs are given in the Appendix.

\subsection{Mapping Network}
\begin{figure}[t]
  \centering
   \includegraphics[width=0.98\linewidth]{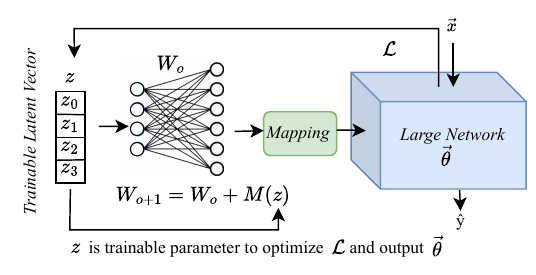}
   \caption{General Architecture for Mapping Networks.}
   \label{fig:genMet}
\end{figure}

Convolutional Neural Networks (CNNs) or in general any deep neural networks exhibit remarkable representational capacity but at the cost of millions of trainable parameters and heavy optimization overhead. 
To address this, we introduce the \emph{Mapping Networks}, as shown in Figure~\ref{fig:genMet}, a meta-learning architecture in which the target network is not directly trained. 
Instead, a trainable latent vector ($z$) and fixed mapping modulated by $z$, generates the target network's parameters, which are used for feed-forward only. 
The result is a substantial reduction in trainable parameters without compromising accuracy or expressivity as proved by the Mapping and Solvability Theorems.

Let the target network whoose training we want to avoid be denoted by $f_{\theta}(\mathbf{x})$ where $\theta = \{W^{(l)}, b^{(l)}\}_{l=1}^{L}$ contains the weights and biases of all the layers, and $\mathbf{x}$ denotes the input image or feature sequence. 
The total number of trainable parameters is $P = \sum_{l=1}^{L} (|W^{(l)}| + |b^{(l)}|)$.


\subsubsection{Trainable Latent Vector}
The latent vector $z$ is made trainable, to allow the model to learn an optimal low-dimensional embedding that best captures the target network's effective parameter distribution. The length of the latent vector is determined in relation to the target network's parameterization and is treated as a tunable hyperparameter to achieve optimal performance. 

\subsubsection{Mapping Network with weight modulation}
\label{sec:modulation}
\begin{figure}[t]
  \centering
   \includegraphics[width=0.98\linewidth, height=230pt]{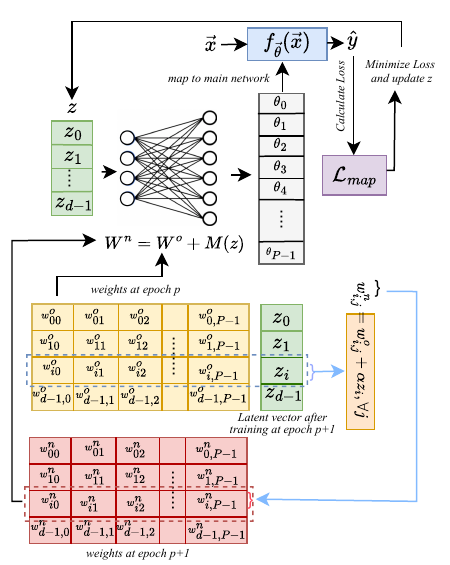}
   \caption{Process of modulation of Mapping weights and training of latent vector z from epoch p to p+1.}
   \label{fig:Modwt}
\end{figure}

The latent vector $z$ is processed by a neural network with fixed non-trainable orthogonally initialized weights, modulated by z. The modulation is introduced as the fixed weights to provide context and prevent the projection being random. The modulation is done by a simple affine transformation of the weights of the mapping network as in Figure~\ref{fig:Modwt}. Let the latent vector($z$) be of dimension $d$, with elements $z_0, z_1,..z_i,..z_{d-1}$ and let $w_{ij}$ be the non-trainable neural ntworks's weights connected to $z_i \forall j=1,2,..,P$, then $w_{ij}$ is modulated as
\begin{equation}
w_{ij} \leftarrow  w_{ij}+ \alpha z_i, \quad \forall j=1,2,..,P
\end{equation}
where $\alpha$ is a small modulation scale.  
Hence, the generated parameters can be given by:
\begin{align}
\mathbf{\hat{\theta}} &= \sigma(\textbf{W}\cdot z + b),
\end{align}
where $\sigma(\cdot)$ denotes activation function. 
The output $\hat{\theta} \in \mathbb{R}^{P}$ represents a flattened, high-dimensional descriptor of the target network's weight space.

\subsubsection{Mapping to Network's Parameters}
The output $\hat{\theta}$ is partitioned and reshaped to match the parameter tensors of each layer of the target network. 
Let $p_l$ and $q_l$ be cumulative indices corresponding to layer $l$'s weights and biases.
Then target network's weights and biases for a layer $l$ can be mapped as,
\begin{align}
W_t^{(l)} &= \mathrm{reshape}\!\big(\mathbf{\hat{\theta}}{[p_l : p_l + |W^{(l)}|]}, \mathrm{shape}(W^{(l)})\big), \\
b_t^{(l)} &= \mathrm{reshape}\!\big(\mathbf{\hat{\theta}}{[q_l : q_l + |b^{(l)}|]}, \mathrm{shape}(b^{(l)})\big).
\end{align}
This deterministic reshaping operation allows $g$ to generate a full set of parameters of the target network from a compact latent representation, without training the network. 

\subsubsection{Target Network for feedforward and Inference}
\label{sec:feedforward}
The Target network then performs standard feed forward and inference:
\begin{equation}
 \hat{y} = \sigma(W_t^T\mathbf{x} + b_t),
\end{equation}
while gradients propagate exclusively through the mapping networks.

\subsubsection{Architecture Add-Ons}

\textbf{Low Rank Decomposition (LRD):  }
The parameter count in fully connected layers can be significantly reduced by applying low-rank decomposition. For a weight matrix $W \in \mathbb{R}^{m \times n}$, LRD approximates it as, $W \approx U V^\top,$ where $U \in \mathbb{R}^{m \times r}$ and $V \in \mathbb{R}^{n \times r}$ with $r \ll \min(m, n)$. 
This reduces the parameter count from $mn$ to $r(m + n)$, which can be substantial for large layers. 
The mapping network then generates the smaller matrices $U$ and $V$ instead of the $W$, which makes our approach memory efficient and scale to even larger target networks.

\textbf{Pruning and Quantization:  }These techniques being completely orthogonal to ours can be easily integrated to Mapping Networks to decrease the inference time and ease deployment on edge devices as shown in Figure~\ref{fig:relwor}.

\subsubsection{Extension to Fine Tuning}

Mapping Networks provide a way to fine tune pre-trained networks by generating modulation vectors $o$ instead of actual parameters. These modulation vectors can be used to tune the pre-trained parameters. Let $W$ denote the pre-trained weights and $W_f$ the fraction of weights to be fine tuned. Now, a pretrained model generally has a huge parameter count, and generating unique modulation elements for each weight will make the mapping network memory inefficient, as the fixed mapping weights though not trained needs to be stored during training. To solve this, let each $o_i$ modulate $L$ weights of $W_f$, then to fine tune $P=|W_f|$ parameters, the mapping network generates $\frac{P}{L}$ modulation elements $(o_1, o_2,...,o_{\frac{P}{L}})$. This will also help to fine tune entire network with very less trainable parameters. To understand the formulation, let's visualize flattened $W_f$ as, $w_{ij}$ with $i=1,2,...,P/L$ and $j=1,2,...,L$.  The modulation occurs as
\begin{equation}
    w_{ij} \leftarrow w_{ij} + \alpha \cdot o_i, \quad \forall j=1,2,...,L
\end{equation}

where $\alpha$ is a small modulation scale. These weights along with the frozen ones are used for feed forward of target network in a similar way as in Section~\ref{sec:feedforward}. Moreover, using Layer wise training (Section~\ref{sec:Layer Wise Training}), we can set different modulation rates ($\alpha's$) for different layers.
\subsection{Mapping Loss}
\label{sec:mappingloss}

To effectively train our Mapping network, the loss function must simultaneously ensure strong downstream task performance and preserve the structural regularity of the parameter manifold, implementing assumptions in our Mapping Theorem. We propose a \textit{Mapping Loss} function:
\begin{equation}
\mathcal{L}_{map} = \mathcal{L}_{\text{task}} + \lambda_{\text{st}}\cdot \mathcal{L}_{\text{stab}} + \lambda_{\text{sm}} \cdot \mathcal{L}_{\text{smooth}} + \lambda_{\text{al}} \cdot \mathcal{L}_{\text{align}},
\end{equation}

where $\lambda_{\text{stab}}, \lambda_{\text{smooth}}, \lambda_{\text{align}}$ are trainable coefficients that control the contribution of each regularization term. These trainable coefficients ensures the mapping network learns to balance task performance and regularization adaptively.

\textbf{Task Loss:  }
The task loss enforces correct predictions for the downstream objective in the target network. For classification tasks, we employ cross-entropy:

\begin{equation}
\mathcal{L}_{\text{task}} = - \sum_i y_i \log \hat{y}_i,
\end{equation}

where $y_i$ is the ground-truth label and $\hat{y}_i$ is the predicted probability output. This term ensures that the generated parameters remain functionally optimal for the target task.\\

\textbf{Stability Loss:  }
Stability loss penalizes large output changes due to small perturbations in the latent vector $z$, this constraint is introduced to enforce the first assumption of our theorem.

\begin{equation}
\mathcal{L}_{\text{stability}} = \mathbb{E} \big[ | f_{\theta'}(z + \epsilon) - f_{\theta'}(z) |_2^2 \big],
\end{equation}

where $f_{\theta'}$ denotes the target network modulated via the mapping $M_\phi(z)$ and $ \epsilon \sim \mathcal{N}(0, \sigma^2 I)$. This term enforces local Lipschitz continuity in the latent space, making the mapping robust to small latent perturbations.  It ensures small latent perturbations do not induce large output deviations.\\

\textbf{Smoothness Loss:  }
To guarantee \textit{smooth differentiability} of the induced parameter manifold, we penalize the Jacobian norm of the mapping $M_\phi(z)$:
\begin{equation}
\mathcal{L}_{\text{smooth}} = \| \nabla_z M_\phi(z) \|_F^2,
\end{equation}
where $\|\cdot\|_F$ denotes the Frobenius norm. This term enforces $C^2$-continuity in the latent-to-parameter transformation, discouraging oscillatory behavior and promoting geometrically consistent transitions in the generated weight space.

\textbf{Alignment Loss:  }
 Maintains compatibility between latent space and mapping weight space, improving generalization.
\begin{equation}
\mathcal{L}_{\text{alignment}} = 1 - \cos\big(z, \overline{W}_{m}),
\end{equation}

where $\overline{W}_{m}$ is the row wise mean of the modulated projection layer weights, and $\cos(\cdot, \cdot)$ denotes cosine similarity. This term aligns the latent vector with the target network’s dominant weight directions.

\subsection{Training}
We partly trained our models on Kaggle's P100 GPU and partly on NVDIA T1000. The datasets MNIST, Fashion MNNIST are taken from Pytorch's datasets. The training strategies used are:
\subsubsection{Single Latent Vector Training (SLVT)}
\label{sec:Full Network Training}
 Here, all the parameters of the target network are approximated by a single trainable latent vector and its modulated mapping weights as in Figure~\ref{fig:nt}. But, with increasing size of target network, the number of non trainable mapping weights will increase, increasing the requirement of system's RAM.
\subsubsection{Layer wise Training (LWT)}
\label{sec:Layer Wise Training}
\begin{figure}[ht]
    \centering
    \begin{subfigure}[b]{0.16\textwidth}
        \centering
        \includegraphics[width=\textwidth, height=100pt]{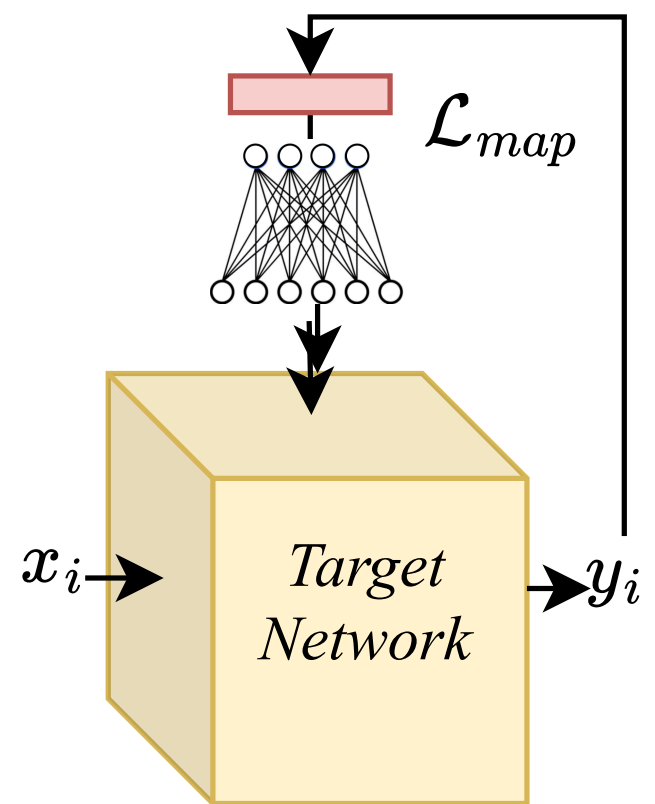}
        \caption{Full Network training by one Latent vector}
        \label{fig:nt}
    \end{subfigure}
    \hfill
    \begin{subfigure}[b]{0.31\textwidth}
        \centering
        \includegraphics[width=\textwidth, height=120pt]{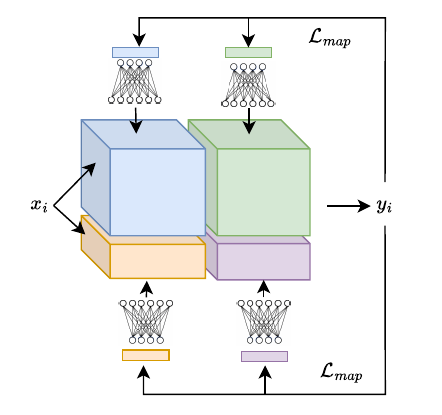}
        \caption{Layer wise training by separate latent vectors}
        \label{fig:complex}
    \end{subfigure}
    
    \caption{Training strategies used for Mapping Network}
    \label{fig:pca_submanifold22}
\end{figure}

To solve the above problem, for large networks having numerous layers, whose parameters might lie in different manifolds, we use separate smaller latent vectors to approximate the parameters of each layer separately as in Figure~\ref{fig:complex}.

\section{Results}
\label{sec:results}

 We have evaluated the performance of our models on various datasets and compared them with baseline methods (Full Results in Appendix). The proposed Mapping Networks Ours* and Ours$\dagger$ represent networks trained by Single Latent Vector and Layer wise training respectively.
\subsection{Mapped CNN Results}
We have tested our Mapping CNN on tasks like image classification, deepfake detection, and image segmentation. For this, we have chosen three baselines CNN1, 2, 3 variants inspired from AlexNet, Le-Net, U-Net (details of architecture in Appendix). 
\subsubsection{Image Classification}
Table~\ref{tab:imageCls}, showcases the test results of our mapping networks on image classification across MNIST, Fashion MNIST datasets. Baselines CNN1 and CNN2, achieve accuracy of 99.32\%, 98.69\% on MNIST, 92.89\%, 90.40\% on FMNIST, respectively with 537,994 and 108,618 parameters. Our mapping networks surpass accuracy of target CNN1 with 2072 parameters only, a 260$\times$ reduction, and almost same accuracy of target CNN2 with 2048 parameters when trained using one latent vector. Furthermore, Ours* surpasses baseline CNN1 with 1024 parameters for FMNIST, achieving a \textbf{525$\times$} reduction. Moreover, layer wise training surpasses both as in Table~\ref{tab:imageCls}, achieving 99.67\% and 94.83\% accuracy. During training of baseline CNN1 on FMNIST, we have got training accuracy 99.10\% but the test accuracy dropped to 92.89\%. On contrast, for the proposed mapping networks with 2072 parameters, this drop was just 1.8\% which is a significant reduction in overfitting.
\begin{table}[h]
\centering
\caption{Image Classification with Mapping CNN}
\label{tab:imageCls}
\begin{tabular}{cccccc} 
\toprule
Method & \# Params & MNIST & FMNIST\\
\midrule
CNN1 & 537,994 & 99.32\% & 92.89\%\\
\midrule
Ours*& 1024 & 98.78\% & \textbf{93.02\%} \\
Ours*& 2072 & \textbf{99.56\%} &\textbf{93.91\%}\\
Ours$\dagger$ & 4078 & \textbf{99.67\%}  &\textbf{94.83\%}\\
\midrule
\midrule
CNN2 & 108,618 & 98.69\% & 90.40\%\\
\midrule
Ours*& 1024 & 97.88\% & 89.49\% \\
Ours*& 2048 & 98.66\% &\textbf{91.88\%}\\
Ours$\dagger$ & 1872 & \textbf{98.98\%}  &\textbf{92.84\%}\\
Ours$\dagger$ & 2688 & \textbf{99.18\%}  &\textbf{93.35\%}\\
\bottomrule
\end{tabular}
\end{table}

\subsubsection{Deepfake Detection}
We have tested our model on a very important and relevant task of deepfake detection of videos on Celeb-DF\cite{li2020celeb} and FF++\cite{rossler2019faceforensics++} dataset. Baseline CNN2 with 108612 trainable parameters gets a test accuracy of 79.03\% on Celeb-DF, whereas our mapping network shows 85.90\% accuracy with just 2048 trainable parameters as in Table~\ref{tab:DeepfakeDet}. A similar improvement is also observed for FF++ dataset. Our Mapping Networks achieve a 5.7\% test accuracy improvement for CNN1 on Celeb-DF using Full Network training strategy. Layer wise training, increases the performance a bit more achieving 86.09\% and 86.28\% accuracy on Celeb-DF and FF++.

\begin{table}[h]
\centering
\caption{Mapping CNN on Deepfake Detection}
\label{tab:DeepfakeDet}
\begin{tabular}{ccccc} 
\toprule
Method & \# Params & Celeb-DF & FF++ \\
\midrule
CNN1 & 537,994 &83.13\% & 82.44\%\\
Ours* & 1024 & \textbf{83.92\%} &  81.11\%\\
Ours* & 2048 & \textbf{88.88\%} &  \textbf{85.23\%}\\
Ours$\dagger$&1956&88.78\%&86.23\%\\
Ours$\dagger$&2792&\textbf{89.98\%}&\textbf{88.05\%}\\
\midrule
CNN2 & 108,618 & 79.03\%& 79.85\%\\
Ours* & 1024 & 78.83\% & 82.78\%  \\
Ours* & \textbf{2048} & \textbf{85.90\%} & \textbf{84.09\%} \\
Ours$\dagger$ & 1872 & 84.54\%  &83.10\%\\
Ours$\dagger$ & 2688 & 86.09\%  &86.28\%\\
\bottomrule
\end{tabular}
\end{table}

\subsubsection{Image Segmentation}
We have also tested our mapping networks on image segmentation task on the cityscapes dataset \cite{cordts2015cityscapes}. Table~\ref{tab:ImageSeg}, shows that our mapping networks with just 8192 parameters achieves a pixel accuracy of 97.92\% and mIoU of 0.4623 and Ours$\dagger$ achieves 97.56\% and 0.48 mIoU, whereas the baseline CNN3 with 1,734,803 parameters achieves a pixel accuracy of 93.21\% and mIoU of 0.4957. Therefore, mapping networks show significant parameter reduction of 211$\times$ while maintaining almost same performance in image segmentation tasks.

\begin{table}[h]
\centering
\caption{Results on Image Segmentation }
\label{tab:ImageSeg}
\begin{tabular}{ccccc} 
\toprule
Method & \# Total &Pixel Acc & Loss & mIoU\\
\midrule
CNN3 & 1,734,803&93.21\%&0.1506& \textbf{0.4957}\\
Ours* & 8192 &97.92\%&0.1233& 0.4623\\
Ours$\dagger$& 9126& 97.56\%&0.1002& 0.4823\\
\bottomrule
\end{tabular}
\end{table}

\subsection{Mapped LSTM Results}
We have tested our Mapping LSTM model on time series analysis on an air pollution dataset (Dataset given in supplementary material). The baseline LSTM model achieves MSE of 0.0035 with 12961 parameters but Mapping Networks surpass it with just 64 parameters, and scales further to 0.00061 with increase in latent size as shown in Table~\ref{tab:TimeSeries}.

\begin{table}[h]
\centering
\caption{Mapping LSTM on Air Pollution dataset}
\label{tab:TimeSeries}
\begin{tabular}{ccccc} 
\toprule
Method & \# Params & MSE Loss\\
\midrule
LSTM & 12961 & 0.0035\\
Ours* & 64 &0.0019\\
Ours* & 2048 &\textbf{0.00061} \\
\bottomrule
\end{tabular}
\end{table}


\begin{table}[h]
\centering
\caption{Fine Tuning ResNet50 via Mapping Networks}
\label{tab:FineTuning}
\begin{tabular}{ccccc} 
\toprule
Method & \# Params & Layers & CDF & FF++\\
\midrule
ResNet50 & 25M & All & 95.23\% & 91.78\%\\
Ours* & 2048 & All& 95.10\% & 91.02\%\\
ResNet50 & 17M & L-4, FC& 91.11\%& 88.03\%\\
Ours* & 1024 & L-4, FC& 92.10\% & 89.23\%\\
\bottomrule
\end{tabular}
\end{table}

\begin{table*}[!htbp]
\centering
\caption{Ablation of Mapping Loss on FashionMNIST dataset}
\label{tab:LossAblation}
\begin{tabular}{ccccccccc} 
\toprule
Method & \# Params & Task Loss & + Stab & + Smooth& + Alli& + Sm + Ali& + Stab + Sm& Full\\
\midrule
CNN2 & 108,618 &90.40\% & -&-&-&-&-&-\\
Ours* & 1024 &87.79\%  & 88.30\% &88.85\% &88.21\%&88.66\% &88.43\% & 89.49\%\\
Ours* & 2048 &87.88\%  & 89.91\% & 90.23\%&90.11\%& 89.86\% & 90.67\%& 91.88\%\\
Ours$\dagger$&1872&89.11\%& 89.56\% & 89.43\%& 89.32\% & 90.47\%& 91.11\%& 92.84\%\\
Ours$\dagger$&2688&91.11\%&91.89\%& 91.50\%& 91.67\%& 92.90\%& 93.63\%& 94.08\%\\
\bottomrule
\end{tabular}
\end{table*}

\begin{table}[h]
\centering
\caption{Robustness Study of Mapping CNN}
\label{tab:ablationModel}
\begin{tabular}{ccccc} 
\toprule
Method & \# Params & MNIST & FMNIST\\
\midrule
CNN2 & 108,618 & 98.69\% & 90.40\%\\
Full DNN & 6,753,104 & 97.12 \% & 90.11\%\\
\midrule
Ours*-- WM & 1024 & 95.62\%  & 86.51\%\\
Ours*-- WM  & 2048 & 96.55\% & 87.66\%\\
\midrule
Ours*& 1024 & 97.88\% & 89.49\% \\
Ours*& 2048 & \textbf{98.66\%} &\textbf{91.88\%}\\
\midrule
LV  + WMAP& 2048 & 97.90\%  &89.30\%\\
LV  + WMAP& 4096 & 98.48\%  &91.93\%\\
\midrule
LV + FullDNN & 543,095 & 96.16\%& 90.11\%\\
LV + FullDNN & 1,629,285 & 97.60\%& 90.67\%\\
\bottomrule
\end{tabular}
\end{table}

\begin{table}[h]
\centering
\caption{Imapct of Add Ons on Mapping Network}
\label{tab:addons}
\begin{tabular}{cccccc} 
\toprule
Method & \# Params & MNIST & FMNIST\\
\midrule
CNN2 & 108,618 & 98.69\% & 90.40\%\\
CNN2 + LRD & 35,914 & 98.12\% & 89.67\%\\
CNN2 + Prune & 10862 & 95.87\% & 87.91\%\\
\midrule
Ours*& 2048 & 98.66\% &91.88\%\\
Ours* + LRD& 1456 & 97.80\% &90.67\%\\
Ours*+ Prune& 2048 & 95.93\%&88.70\%\\
\midrule
Ours$\dagger$ & 2688 & 99.28\%  &94.08\%\\
Ours$\dagger$+LRD & 2688 & 98.81\%  &93.55\%\\
Ours$\dagger$ + Prune& 2688 & 97.15\%  &91.79\%\\
\bottomrule
\end{tabular}
\end{table}

\subsection{Fine-Tuning a Model}
 Table~\ref{tab:FineTuning} presents the results of fine-tuning ResNet50 on deepfake detection. The results indicate that our mapping networks can effectively adapt pre-trained models to new tasks with a significantly reduced of trainable parameters while achieving competitive accuracy 95.10\% and 91.02\% using $L=250$. Variation of results with varying $L$ is shown in appendix.

\subsection{Ablation of Mapping Networks}
Table~\ref{tab:ablationModel} presents an ablation study of our Mapping CNN on FashionMNIST dataset. Full DNN represents that the latent vector is not trainable but the mapping weights are trainable via gradient descent. Next, Ours*–WM denotes that the mapping weights are fixed and not modulated. LV + WMAP means the mapping weights are modulated by another set of trainable parameters not the latent vector. Finally, in LV + FullDNN both the latent vector and mapping weights are trainable, where latent vector lengths of 5 and 15 correspond to $5 \times 108618+5$ and $15 \times 108618+15$ trainable parameters respectively. Here, increasing latent vector further explodes parameter count without helping much.
Among all others, Mapping Networks (Ours*) achieves best results, even using separate trainable parameters in LV + WMAP does not improve the performance. The 2-4\% accuracy increment in Ours* over Ours* - WM, highlights the importance of weight modulation. Furthermore, outperforming both FullDNN and LV + FullDNN shows that making the mapping weights fully trainable does not help, but increases overfitting. Thus, modulating the mapping weights provides an effective trade-off between underfitting and overfitting.
\subsection{Ablation of Mapping Loss}
Table~\ref{tab:LossAblation} presents an ablation study of our Mapping CNN using various combinations of loss components in the Mapping loss function on the FashionMNIST. The results indicate that different configurations of the mapping loss yield increasing accuracy as various components are added. From the table its quite clear that the stability and smoothness component contribute slightly more than alignment one. The overall accuracy increment of 2-3\% from task loss to mapping loss, proves its importance.

\subsection{Impact of Add-Ons}
Table~\ref{tab:addons} shows the effect of 90\% unstructured pruning and low rank decomposition (rank 16) of fully connected layers in baseline and Mapping Networks.  Ours* and Ours$\dagger$ has the same number of non trainable inference parameters as target CNN2 for inference. The effect of pruning and LRD on CNN2, Ours* and Ours$\dagger$ are quite similar, showing those techniques can be effectively combined with Mapping Networks as well. So, Ours*/Ours$\dagger$+LRD and Pruning effectively decreases both training and inference parameters and time, enabling it to get the perfect spot in Figure~\ref{fig:relwor}.
\section{Conclusion}
Mapping Networks achieve comparable to better performance on presented tasks with 200$\times$ to 500$\times$ less trainable parameters while increasing expressivity, reducing overfitting and training time of the model. However, for large target networks SLVT becomes slightly memory expensive. This is solved by layer wise training which reduces the memory requirement almost 10 times. For fine tuning, this is jointly solved by layer wise training and increasing count of fine tuned weights per modulation element, enabling us to extend Mapping Networks to LLMs and LVMs in future. 
\newpage
\clearpage
{
    \small
    \bibliographystyle{ieeenat_fullname}
    \bibliography{ref}
}
\end{document}